\documentclass[conference]{IEEEtran}
\IEEEoverridecommandlockouts

\usepackage{fancyhdr} 
\fancypagestyle{firstpageheader}{
  \fancyhf{} 
  \fancyhead[C]{\footnotesize \color{blue} NOTICE: This arXiv version is an extended version of our paper that has been published in IEEE Micro Journal.} 
}

\usepackage{amsmath,amssymb,amsfonts}
\usepackage{algorithmic}
\usepackage{graphicx}
\usepackage{textcomp}

\usepackage[hyphens]{url}
\usepackage{caption}
\usepackage{subcaption}
\usepackage[hyphens]{url}
\usepackage{graphicx}
\usepackage{float}
\usepackage{graphbox}

\usepackage{soul}
\usepackage[utf8]{inputenc}
\usepackage{graphicx}
\usepackage{booktabs}
\usepackage[square,numbers,sort&compress]{natbib}

\usepackage{graphicx}

\usepackage[colorlinks=true,
citecolor=blue,
filecolor=black,
linkcolor=blue,
urlcolor=blue]{hyperref}

\usepackage[ruled,noend]{algorithm2e}

\SetCommentSty{mycommfont}

\usepackage{here}

\usepackage{amsmath,amssymb,amsfonts,amsbsy,amsfonts,latexsym}
\usepackage{multirow}
\usepackage{makecell}
\usepackage[labelfont=bf,textfont=it,belowskip=0pt,aboveskip=5pt,tableposition=top]{caption}
\usepackage{xcolor}
\usepackage{colortbl}

\definecolor{colorA}{RGB}{189,201,225}
\definecolor{colorB}{RGB}{103,169,207}
\definecolor{colorC}{RGB}{ 28,144,153}
\definecolor{colorD}{RGB}{  1,108, 89}

\newcolumntype{R}{>{\columncolor{gray!40}}r}
\newcolumntype{L}{>{\columncolor{gray!40}}l}
\newcolumntype{C}{>{\columncolor{gray!40}}c}

\usepackage{tabularx,colortbl,xcolor}
\usepackage{multirow}
\usepackage[normalem]{ulem}
\useunder{\uline}{\ul}{}

\usepackage{enumitem}

\usepackage{xparse}

\DeclareGraphicsExtensions{.pdf,.png}

\SetKwInput{KwInput}{Input}

\usepackage{longtable}
\usepackage{pgfplots}
\usepackage{outlines}

\newcommand{\fref}[1]{Figure~\ref{#1}}

\usepackage{tikz}
\usetikzlibrary{backgrounds}
\makeatletter

\tikzset{%
  fancy quotes/.style={
    text width=\fq@width pt,
    align=justify,
    inner sep=0.1em,
    anchor=north west,
    minimum width=\linewidth,
  },
  fancy quotes width/.initial={.1\linewidth},
  fancy quotes marks/.style={
    scale=1,
    text=black,
    inner xsep=6pt,
    inner ysep=0pt,
  },
  fancy quotes opening/.style={
    fancy quotes marks,
  },
  fancy quotes closing/.style={
    fancy quotes marks,
  },
  fancy quotes background/.style={
    show background rectangle,
    inner frame xsep=0pt,
    background rectangle/.style={
      fill=gray!25,
      rounded corners,
    },
  }
}

\tikzset{
  plain text box/.style={
    align=justify,
    text width=\linewidth - 2*10pt, 
    inner xsep=10pt, 
    inner ysep=10pt, 
    fill=gray!25, 
    rounded corners 
  }
}

\newenvironment{plaintxtbox}{%
  \par\noindent\begin{tikzpicture}%
  \node[plain text box] (txt) \bgroup%
}{%
  \egroup;%
  \end{tikzpicture}\par%
}

\def\BibTeX{{\rm B\kern-.05em{\sc i\kern-.025em b}\kern-.08em
    T\kern-.1667em\lower.7ex\hbox{E}\kern-.125emX}}
\begin{document}

\title{AI and Memory Wall}

\author{
Amir Gholami$^{12}$
Zhewei Yao$^{1}$ 
Sehoon Kim$^{1}$ 
Coleman Hooper$^{1}$ 
Michael W. Mahoney$^{1,2,3}$, Kurt Keutzer$^{1}$\\
$^{1}$University of California, Berkeley,  $^{2}$ICSI,  $^{3}$LBNL\\
{\tt\small \{amirgh, zheweiy, sehoonkim, chooper, mahoneymw, keutzer\}@berkeley.edu},
}

\author{
\IEEEauthorblockN{
Amir Gholami$^{1,2}$
Zhewei Yao$^{1}$
Sehoon Kim$^{1}$
Coleman Hooper$^{1}$
Michael W. Mahoney$^{1,2,3}$
Kurt Keutzer$^{1}$
}
\vspace{0.1in}
\IEEEauthorblockA{
$^1$University of California, Berkeley \enspace $^2$ICSI \enspace $^3$LBNL}
}

\maketitle
\thispagestyle{firstpageheader}
\pagestyle{plain}

\begin{abstract}
The availability of unprecedented unsupervised training data, along with neural scaling laws, has
resulted in an unprecedented surge in model size and compute requirements for serving/training LLMs.
However, the main performance bottleneck is increasingly shifting to memory bandwidth.
Over the past 20 years, peak server hardware FLOPS has been scaling at 3.0$\times$/2yrs, outpacing the growth of DRAM and interconnect bandwidth, which have only scaled at 1.6 and 1.4 times every 2 years, respectively.
This disparity has made memory, rather than compute, the primary bottleneck in AI applications, particularly in serving. 
Here, we analyze encoder and decoder Transformer models and show how memory bandwidth can become the dominant bottleneck for decoder models.
We argue for a redesign in model architecture, training, and deployment strategies to overcome this memory limitation.
\end{abstract}


\section{Introduction}

The amount of compute needed to train Large Language Models (LLMs) has recently been growing at a rate of 750$\times$/2yrs. This exponential trend has been the main driver for AI accelerators that focus on increasing the peak compute power of hardware, often at the expense of simplifying other parts such as memory~hierarchy.

However, these trends miss an emerging challenge with training and serving AI models: memory and communication bottlenecks.
In fact, several AI applications are becoming bottlenecked by intra/inter-chip and communication across/to AI accelerators, rather than compute.
This is not a new phenomena, and several works in the past observed and warned
about this issue. One of the earliest observations of this dates back to 1990 when
Ousterhout concluded the following after analyzing the factors impacting operating system's performance~\cite{ousterhout1990aren}:

\vspace{5mm}
\begin{plaintxtbox}
``The first hardware-related issue is memory bandwidth: the benchmarks suggest that it is not keeping up with CPU speed ... If memory bandwidth does not improve dramatically in future machines, some classes of applications may be limited by memory performance."
\end{plaintxtbox}
\vspace{1mm}

Later in 1995, William Wulf and Sally Mckee further echoed this prediction and coined the term ``memory wall''.
Their argument for this followed a simple but elegant reasoning.
The time to complete an operation is dependent on how fast we can perform
the arithmetic as well as how fast we can feed data to the arithmetic units of
hardware.\footnote{Just for reference a better way to analyze this is through arithmetic intensity proposed by Sammuel Williams~\cite{williams2009roofline}  which will be discussed in Sec.\ref{sec:workload-analysis}.}
In the simplest case, the data is either
available in the cache, or needs to be fetched from DRAM. With this assumption, even if
$80\%$ of data is readily available in cache, and only $20\%$ needs to be fetched from DRAM,
the time to complete the operation will be completely limited by DRAM if it takes
more than 5 cycles to fetch the $20\%$ cache-miss data from it.
This means that no matter how fast the hardware could perform arithmetics per second, the problem will be entirely limited by DRAM bandwidth. They predicted that the diverging speed of improvement
of how fast computations can be performed versus how fast data can be fetched is going to create a ``memory wall'' issue~\cite{wulf1995hitting,memory-wall-reflections}.
Based on this they concluded:

\vspace{5mm}
\begin{plaintxtbox}
``Each is improving exponentially, but the exponent for microprocessors is substantially larger than that for DRAMs. The difference between diverging exponentials also grows exponentially."
\end{plaintxtbox}
\vspace{1mm}

Several later works also reported a similar observation~\cite{hennessy2011computer,patterson1997case,memory-wall-reflections,sites1996architects, wilkes1995memory, mccalpin2006stream}. 

In this work, we re-examined this trend by studying more recent data, with a particular focus on hardware used to train AI models, as well as the characteristics
of the computations used for training/serving them. 30 years after, the above observations and predictions could not be further correct. 
Despite many innovations in memory technology, the trend shows that the ``memory wall'' is increasingly becoming the dominant bottleneck for a range of AI tasks. 

\begin{figure}[!t]
  \centering
  \hspace{-2mm}\includegraphics[width=0.495\textwidth]{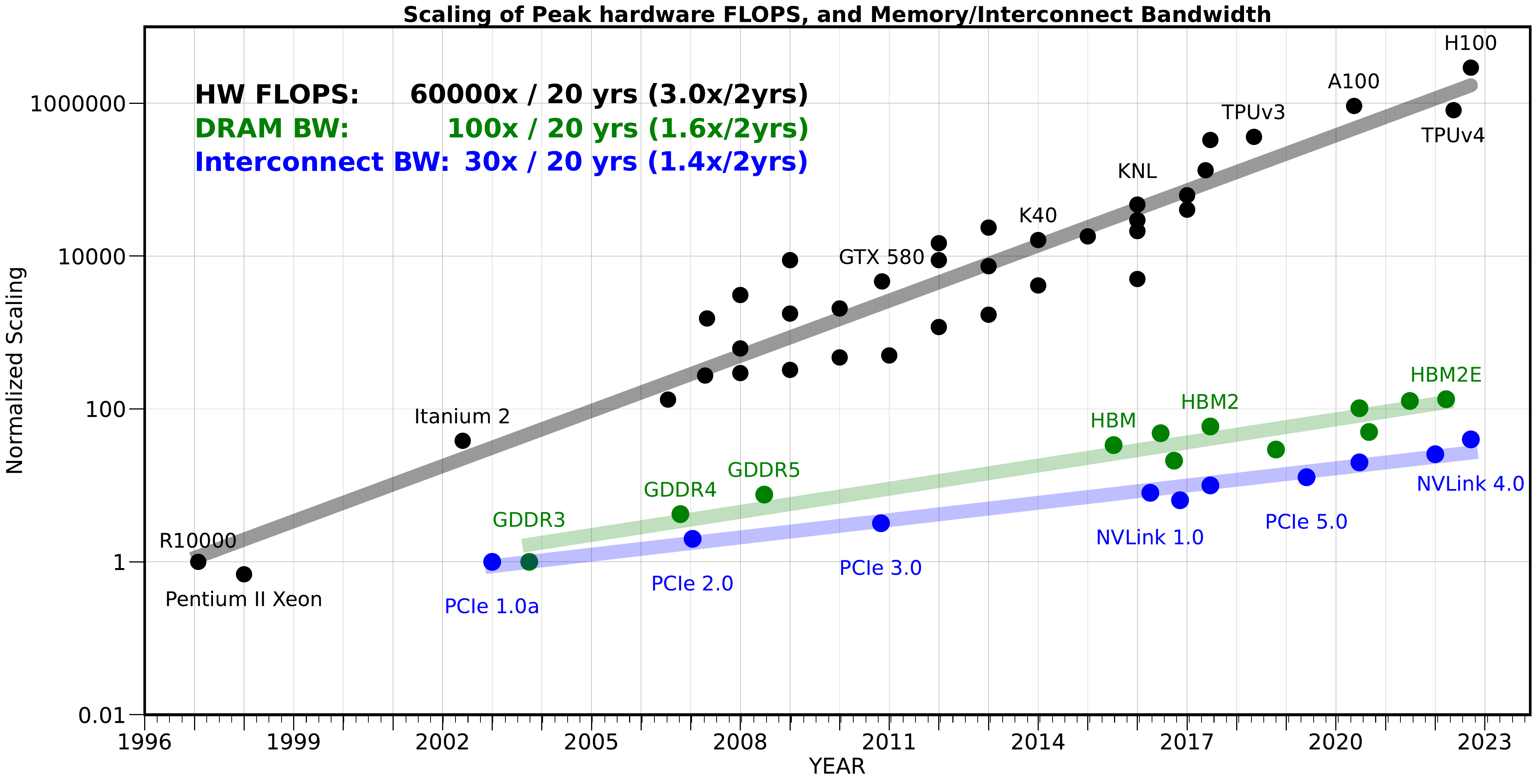}
  \caption{The scaling of the bandwidth of different generations of interconnections and memory, as well as the Peak FLOPS. As can be seen, the bandwidth is increasing very slowly. We are normalizing hardware peak FLOPS with the R10000 system, as it was used to report the cost of training LeNet-5 \cite{lecun1998gradient}.}
      \label{fig:hw_scaling}
\end{figure}

\begin{figure*}[!ht]
    \centering
    \begin{subfigure}{0.49\textwidth}
        \centering
        \includegraphics[width=\linewidth]{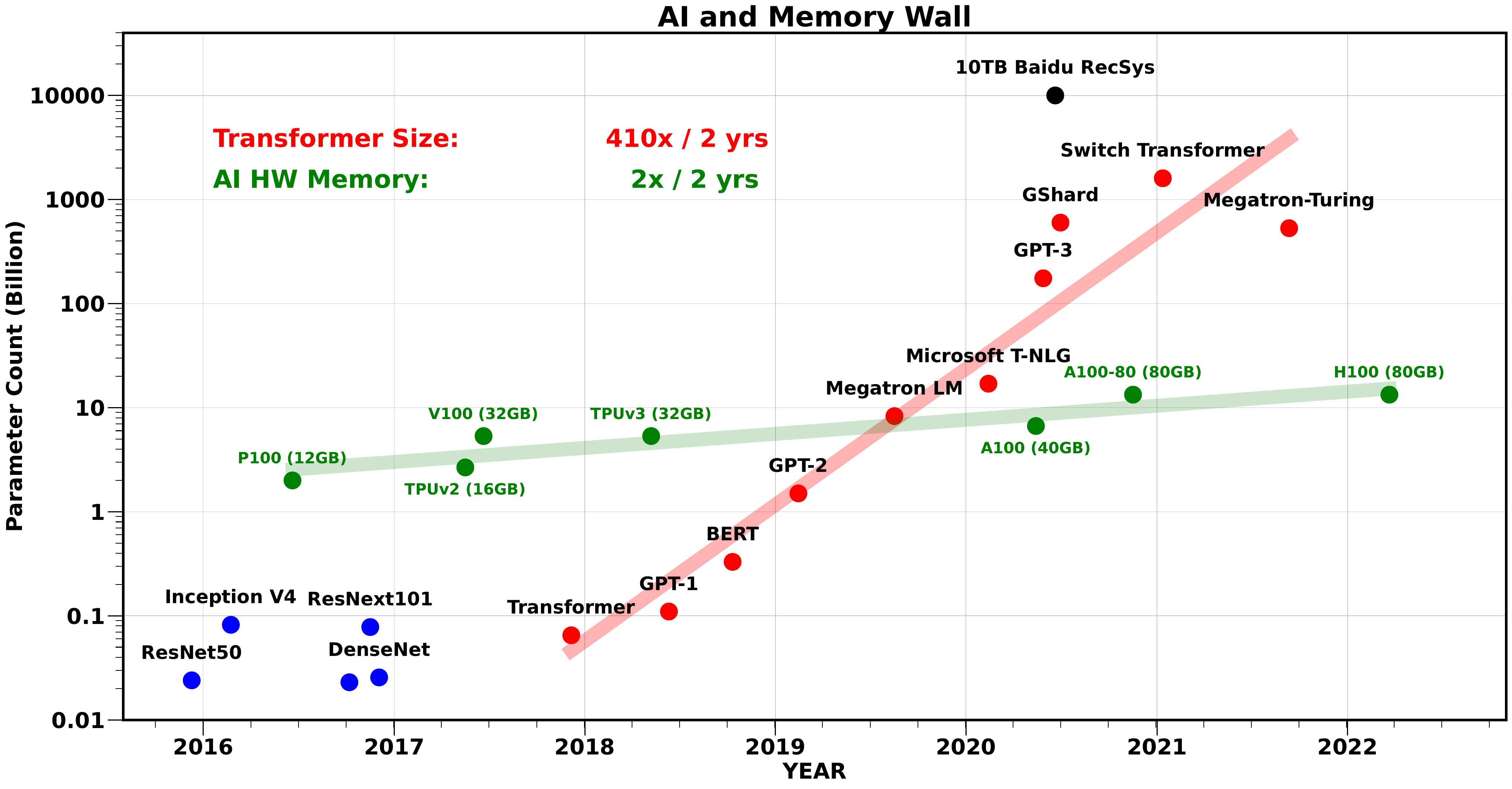}
        \caption{}
        \label{fig:sub1}
    \end{subfigure}%
    \hfill
    \begin{subfigure}{0.49\textwidth}
        \centering
        \includegraphics[width=\linewidth]{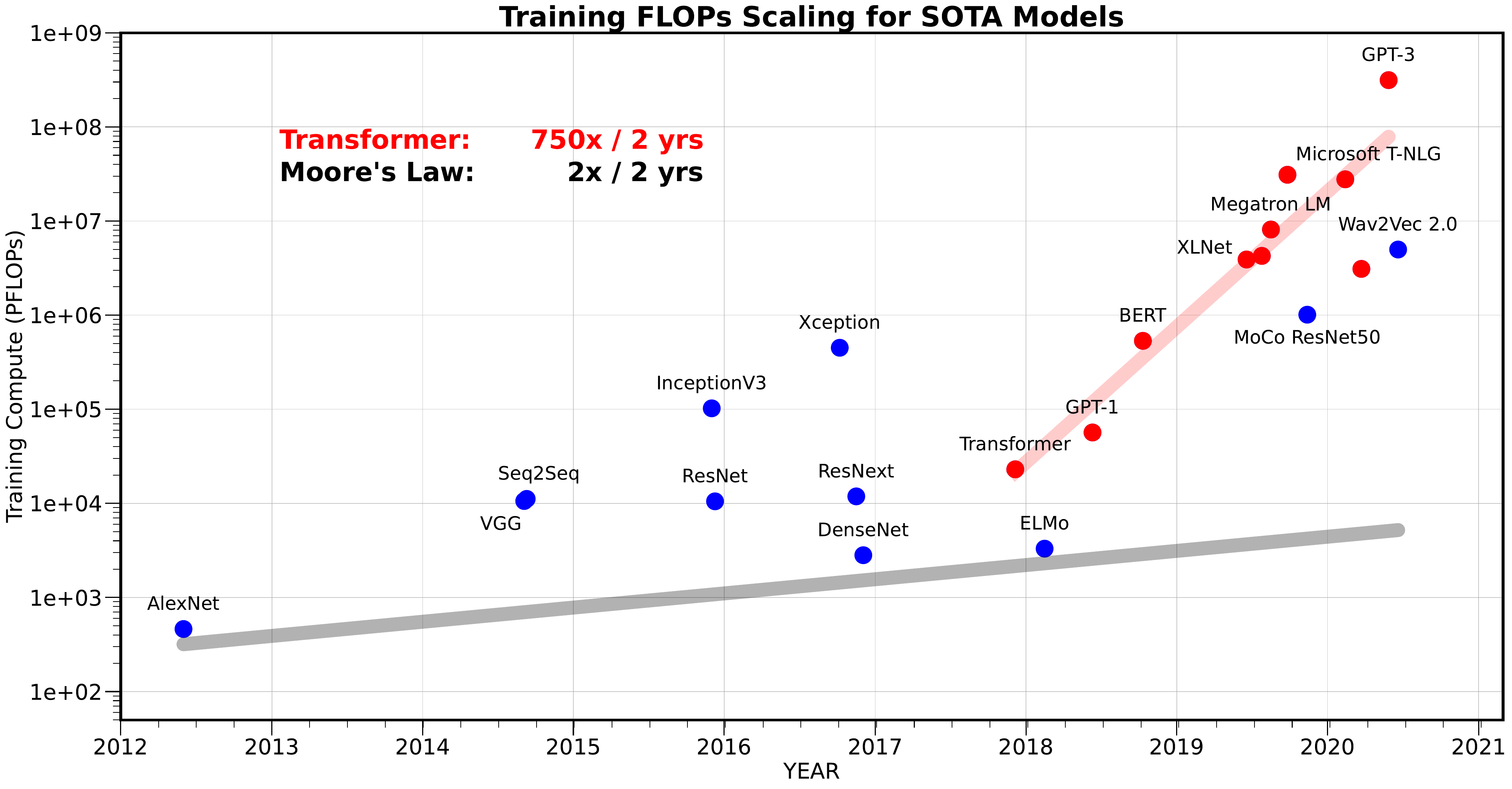}
        \caption{}
        \label{fig:sub2}
    \end{subfigure}
  \caption{(a) The evolution of the number of parameters of state-of-the-art (SOTA) models over the years, along with the AI accelerator memory capacity (green dots). The number of parameters in large Transformer models has been exponentially increasing with a factor of 410$\times$ every two years, while the single GPU memory has only been scaled at a rate of 2$\times$ every 2 years.  The growth rate for the Transformer models is calculated by only considering the non-recommendation system models (red circles), and the GPU memory is plotted by dividing the corresponding memory size by 6 as an approximate upper bound for the largest model that can be trained with the corresponding capacity.
  (b) The amount of compute, measured in Peta FLOPs, needed to train SOTA models, for different computer vision (CV), natural language processing (NLP), and Speech models, along with the different scaling of Transformer models (750$\times$/2yrs).\protect\footnotemark
  }
      \label{fig:ai_and_compute_size}
      \vspace{-2mm}
\end{figure*}
We first start by analyzing how peak compute of server-grade AI hardware has changed since 1998 when Yann Lecun trained the famous Lenet-5 model on MNIST data~\cite{lecun1998gradient}.
We can see that the peak compute of the hardware has increased by 60,000$\times$ over the past 20 years, as opposed to a 100$\times$ increase for DRAM or a 30$\times$ increase for interconnect bandwidth.

The memory wall problem involves both the limited capacity, the bandwidth of memory transfer, as well as its latency (which has been even harder to improve~\cite{patterson2004latency} than bandwidth). This entails different levels of memory data transfer. For example, data transfer between compute logic and on-chip memory, or between compute logic and DRAM memory, or across different processors on different sockets. In all these cases, the capacity and the speed of data transfer has been significantly lagging behind hardware compute capabilities.


Now, if we study the trend of recent AI models, and in particular LLMs, 
we notice that practitioners, motivated by neural scaling law~\cite{hoffmann2022training}, have been
scaling the amount of data, model size, and 
compute needed to train recent models at unprecedented levels.
Even though compute / floating-point operations (FLOPs)\footnote{Please note that we use FLOPs with lowercase s to denote the number of floating point operations
needed to perform a task, and FLOPS with capital S to denote the rate at which a given hardware can perform floating point operations per second.}
needed to train these recent models has increased
by a factor of 750$\times$/2yrs in the 2018-2022 time frame (see \fref{fig:ai_and_compute_size}), compute is not necessarily the bottleneck, especially for model serving.

First, the LLM sizes have scaled at a rate of 410$\times$/2yrs in that time frame, exceeding memory available on single chip. 
One might hope that we can use distributed-memory parallelism by scaling-out the training/serving to multiple accelerators to avoid the single hardware’s limited memory capacity and bandwidth. However, distributing the work over multiple processes can also face the memory wall problem: the communication bottleneck of moving data between neural network (NN) accelerators, which is even slower and less efficient than on-chip data movement. Similar to the single system memory case, we have not been able to overcome the technological challenges to scale the network bandwidth.

Second, even when the model fits within a single chip, intra-chip memory transfers
from/to registers, L2 cache, global memory, etc. are still increasingly becoming the bottleneck. Thanks
to the recent advancements in specialized compute units, such as Tensor cores, the arithmetic operations for a large
set of computations can finish in a few cycles. Therefore, to keep these arithmetic units utilized at all times
one needs to rapidly feed them large amounts of data, and that is where the chip memory bandwidth becomes the bottleneck.

As one can see in \fref{fig:hw_scaling}, over the past 20 years, peak server hardware FLOPS has been scaling at 3.0$\times$/2yrs, outpacing the growth of DRAM and interconnect bandwidth, which have only scaled at 1.6 and 1.4 times every 2 years, respectively.
This disparity has made memory, rather than compute, increasingly become a bottleneck, even for cases when the model can fit within a single chip.

Next, we perform a detailed case study for Transformers which helps better showcase the interplay between FLOPs, Memory Operations (MOPs), and end-to-end
runtime by considering common models used today.

\footnotetext{We are specifically not including the cost of training reinforcement learning models in this figure, as the training cost is mostly related to the simulation environment, and there is currently no consensus on a standard simulation environment. Also note that we report the PFLOPs 
required to train each model to avoid using any approximation for hardware deployment utilization, as the latter depends on the specific library and the hardware used. Finally, all the rates in this document have been computed by solving a linear regression to fit the data shown in each graph.}

\section{Case Study}
\label{subsec:model-analysis}
In this section, we first outline the run-time characteristics and the performance bottleneck associated with Transformer inference. 
We examine two different variations of the Transformer architecture: the encoder architecture (e.g., BERT~\cite{devlin2018bert}), which concurrently processes all tokens, and the decoder architecture (e.g., GPT~\cite{radford2019language,brown2020language}), which runs auto-regressively to process and generate one token at each iteration.

\subsection{\textbf{Arithmetic Intensity}}
\label{sec:workload-analysis}
A popular method for measuring performance bottlenecks is to
compute the total number of FLOPs required to compute the Transformer encoder-only and decoder-only models.
However, this metric in isolation can be very misleading.
Instead, one needs to study the arithmetic intensity of the operations involved.
Arithmetic intensity is the number of FLOPs that can be performed per byte loaded from memory.
It can be computed by dividing the total number of FLOPs by the total number of bytes accessed (also referred to as MOPs, or memory operations)~\cite{williams2009roofline}\footnote{Here, we are assuming that the local memories are large enough to hold both matrices entirely in memory for a given operation, and that the computed arithmetic intensity values therefore serve as an upper bound for the achievable data reuse. We are also counting the multiplication and addition from a MAC operation separately when computing FLOPs.}:
\begin{equation}
    \text{Arithmetic Intensity} = \frac{\text{\# FLOPs}}{\text{\# MOPs}} .
\end{equation}

To illustrate the importance of considering Arithmetic Intensity, we studied BERT-Base and BERT-Large \cite{devlin2018bert}, along with GPT-2 \cite{radford2019language}.
The first two are encoder models, 
which involve matrix-matrix operations for their inference, 
and the latter is a decoder/auto-regressive model, 
where its inference involves repeated matrix-vector multiplications.

\subsection{\textbf{Profiling}}
\begin{figure*}[!ht]
    \centering
    \begin{subfigure}{0.49\textwidth}
        \centering
        \includegraphics[width=\linewidth]{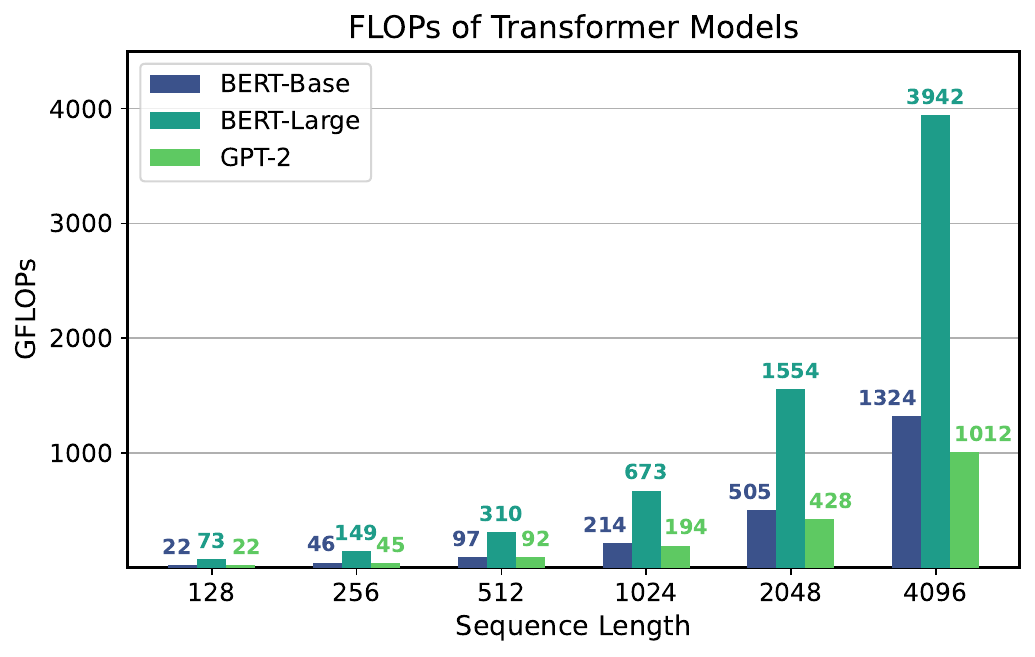}
        \caption{}
        \label{fig:sub1}
    \end{subfigure}%
    \hfill
    \begin{subfigure}{0.49\textwidth}
        \centering
        \includegraphics[width=\linewidth]{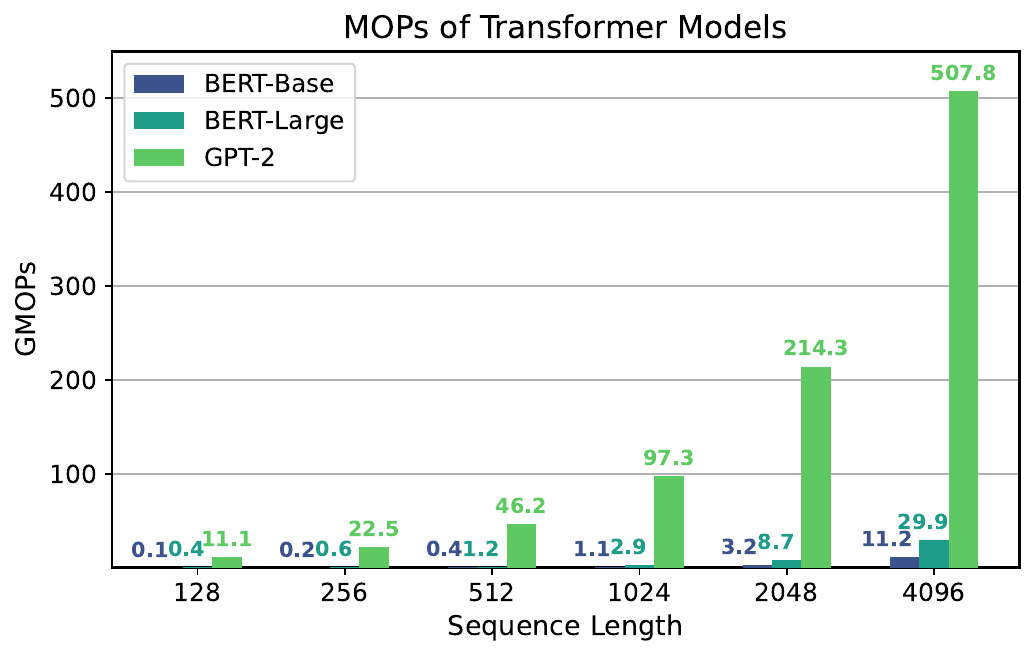}
        \caption{}
        \label{fig:sub2}
    \end{subfigure}
    
    \vspace{1em}
    
    \begin{subfigure}{0.49\textwidth}
        \centering
        \includegraphics[width=\linewidth]{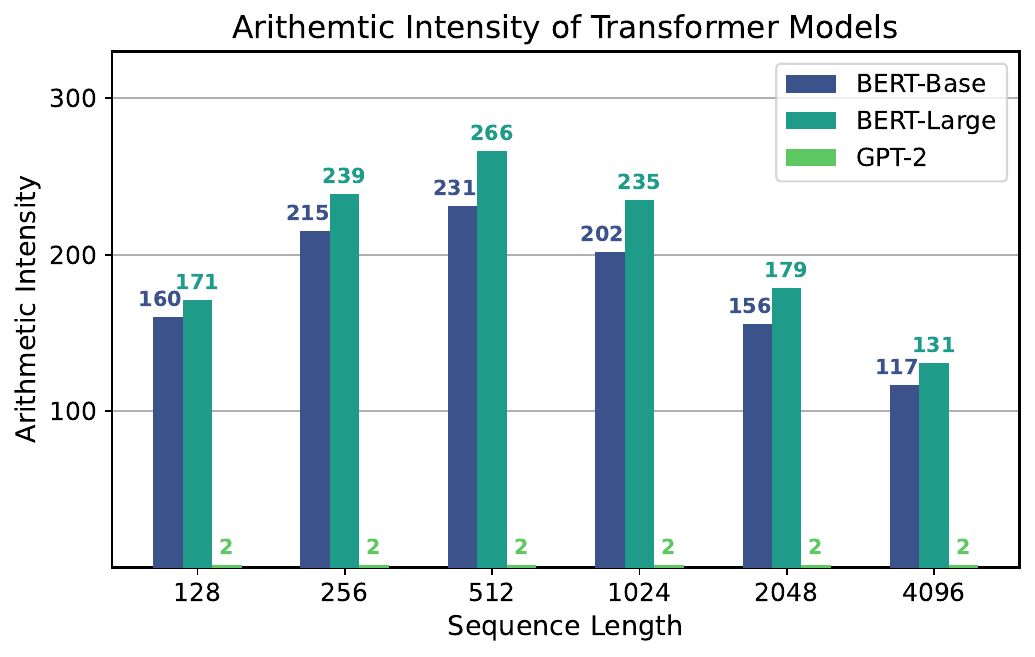}
        \caption{}
        \label{fig:sub3}
    \end{subfigure}%
    \hfill
    \begin{subfigure}{0.49\textwidth}
        \centering
        \includegraphics[width=\linewidth]{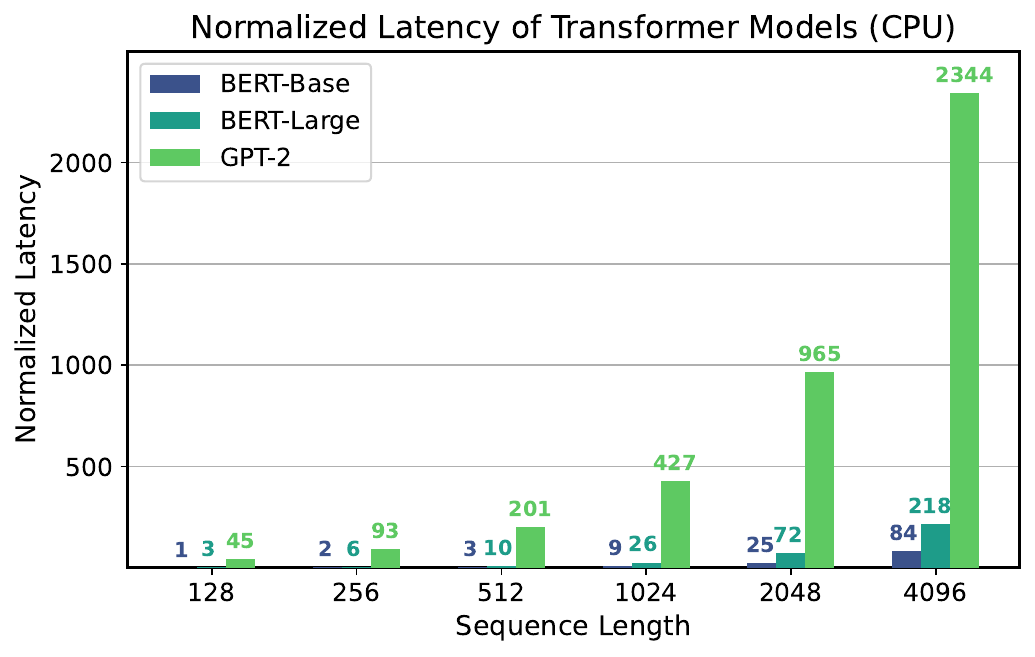}
        \caption{}
        \label{fig:sub4}
    \end{subfigure}
  \caption{Profiling results for BERT-Base, BERT-Large, and GPT-2 models for processing/generating
  different sequence lengths with batch size 1.
  (a) total inference FLOPs: 
  notice how encoder models have higher FLOPs;
  (b) total inference Memory Operations (MOPs): 
  notice how the decoder GPT model has orders of magnitude more MOPs due to its matrix-vector type operations vs matrix-matrix operations in encoder models;
  (c) arithmetic intensity: 
  notice how GPT-2 has orders of magnitude smaller arithmetic intensity, which makes it very challenging to effectively utilize a given hardware's compute units;
  (d) end-to-end latency of the different models normalized to the BERT-Base model for processing an input sequence length of 128: 
  notice how the decoder model's runtime is the slowest, despite having smaller FLOPs. See~\cite{kim2023full} for more details.
  }
\label{fig:all_profiling}
\end{figure*}

To analyze the bottlenecks in Transformer workloads on commodity hardware, we profiled Transformer inference on an Intel Gold 6242 CPU.
\fref{fig:all_profiling} shows the total FLOPs, MOPs, Arithmetic Intensity, and the final latency of these models for different sequence lengths.\footnote{
We assumed that all model parameters and activations are stored in 8-bit precision, and batch
size of 1.
In the case of the decoder model, we measured the total amount of the FLOPs and MOPs needed to iteratively generate the full sequence of the given length.} 
It is evident that the GPT-2 latency is significantly longer than the latency for either BERT-Base or BERT-Large for each sequence length, even though BERT-Base and GPT-2 have largely the same model configuration and end-to-end FLOPs (as is depicted in \fref{fig:all_profiling}a).
This is due to the higher memory operations and lower arithmetic intensity of matrix-vector operations inherent in the auto-regressive inference of GPT (see \fref{fig:all_profiling}c).
A model with higher arithmetic intensity can run faster with the same or possibly even more FLOPs than a model with lower arithmetic intensity. 
This clearly shows how the memory wall can become a major bottleneck for decoder models (at low batch sizes) and not compute.\footnote{
Note that this may not apply to all kinds of applications of decoder models such as in summarizing long documents where the inherent operations for processing the input prompt are matrix-matrix operations. Other cases include large batch size inference which effectively includes matrix-matrix operations.}


\section{Promising Solutions for Breaking the Wall}

\textbf{“No exponential can continue forever,”}~\cite{moore2003no} and delaying an exponential scaling at the rate of 410$\times$/2yrs is not going to be feasible for long, even for large hyperscalar companies. This, coupled with the increasing gap between compute and bandwidth capability, will soon make it very challenging to train larger models, as the cost will be exponentially higher.

To continue the innovations and break the memory wall, we 
need to rethink the design of AI models. There are several 
issues here. First, the current methods for designing AI 
models are mostly ad-hoc, and/or involve very simple scaling 
rules. For instance, recent large Transformer models~\cite{touvron2023llama,brown2020language,jiang2023mistral} are 
mostly just a scaled version of almost the same base 
architecture proposed in the original BERT 
model~\cite{devlin2018bert}. Second, we need to design more 
data-efficient methods for training AI models. Current NNs 
require a huge amount of training data and hundreds of 
thousands of iterations to learn, which is very inefficient. 
Some might note that it is also different from how human 
brains learn, which often only requires very few examples per 
concept/class. Third, the current optimization and training 
methods need a lot of hyperparameter tuning (such as 
learning rate, momentum, etc.), which often results in 
hundreds of trial and error sweeps to find the right hyperparameter setting 
to train a model successfully. As such, the training cost 
reported in \fref{fig:ai_and_compute_size} (b) is only a 
lower bound of the actual overhead, and the true cost is 
typically much higher. Fourth, the prohibitive size of the 
state-of-the-art models makes their deployment for inference 
very challenging. This is not just restricted to models such 
as GPT-3. In fact, deploying large recommendation systems 
that are used by 
hyperscalar companies is a major challenge. Finally, the 
design of hardware accelerators has been mainly focused on 
increasing peak compute with relatively less attention on 
improving memory-bound workloads. This has made it difficult 
both to train large models, as well as to explore 
alternative models, such as Graph NNs which are often 
bandwidth-bound and cannot efficiently utilize current 
accelerators.

All of these issues are fundamental problems in machine learning. Here, we briefly discuss recent research (including some of our own) that has targeted the last three items.

\subsection{Efficient Training Algorithms}

One of the main challenges with training NN models is the need for brute-force hyperparameter tuning. 
This includes finding the learning rate, its annealing schedule, the number of iterations needed to converge, etc. 
This adds (much) more overhead for training SOTA models. Many of these problems arise from the first-order SGD methods used for training. 
While SGD variants are easy to implement, they are not robust to hyperparameter tuning, and are very hard to tune for new models, for which the right set of hyperparameters is unknown.
One promising approach to address this is to 
use second-order stochastic optimization methods~\cite{yao2021adahessian}. 
These methods are typically more robust to hyperparameter tuning, and they can achieve SOTA~\cite{yao2021adahessian}. However, current methods have 3–4$\times$ higher memory footprint, which needs to be addressed~\cite{yao2021adahessian}. 
A promising line of work for that is the Zero framework from Microsoft, which showed how one can train 8$\times$ bigger models with the same memory capacity by removing/sharding redundant optimization state variables~\cite{rajbhandari2020zero,bottou2018optimization}.
If the overhead of these higher-order methods could be addressed, then they can significantly reduce the total cost of training large models.

\begin{figure}[!ht]
  \centering
  \includegraphics[width=0.85\columnwidth]{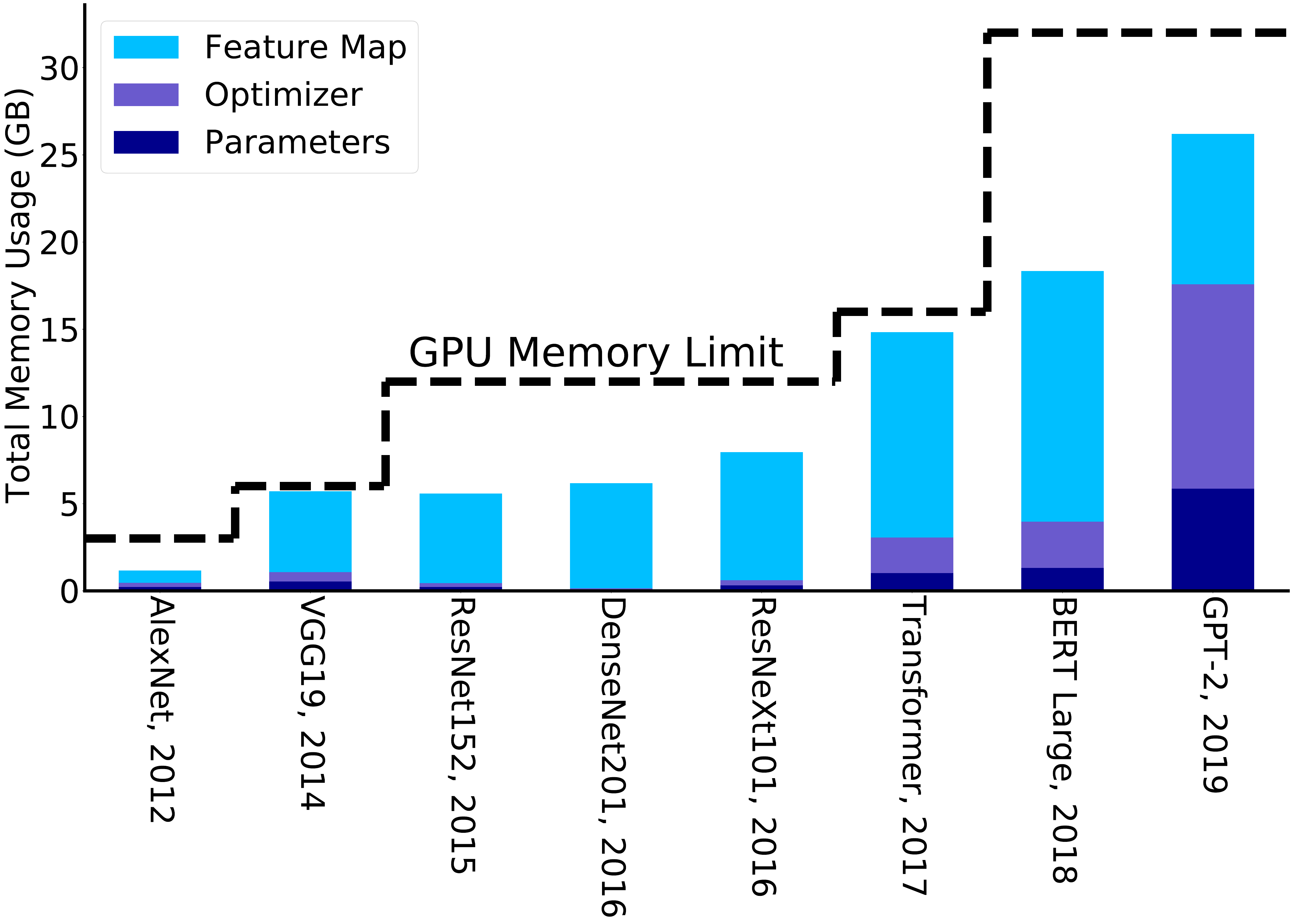}
  \caption{The amount of memory required to train different NN models. Here, the optimizer used for CV models is SGD+Momentum, and for NLP models it is ADAM. There is an interesting trend in discovering/designing new models, based on the available GPU memory size. Every time the GPU memory capacity is increased, data scientists have designed newer models. As such, breaking this so-called GPU memory wall could further allow new innovations. See~\cite{jain2020checkmate} for more details on checkpointing.
  }
      \label{fig:gpu_memory_wall}
\end{figure}
Another promising approach includes reducing the memory 
footprint and increasing the data locality of optimization 
algorithms, at the expense of performing more computations.
A notable example of this in numerical linear algebra is the family of
communication avoiding algorithms~\cite{ballard2011minimizing}.
One example of optimizing for memory for NN training is 
rematerialization, where we only store/checkpoint a subset of 
activations during the forward pass, instead of saving all 
activations.  This reduces the feature map’s memory footprint, as 
shown in \fref{fig:gpu_memory_wall}. The rest of the 
activations could then be recomputed when 
needed~\cite{jain2020checkmate,korthikanti2023reducing}. 
Even though this will increase compute, one can 
significantly reduce the memory footprint by up to 
5$\times$~\cite{jain2020checkmate}, with just 20\% more 
compute. This can also allow practitioners to train large 
models on single-chip memory rather than utilize distributed 
training, which is often difficult to set up (outside of major 
hyperscaler companies) and is hard to debug (for
non-expert developers).
Interestingly, traditional trends show that new NN model architectures
have been developed based on what researchers have access to within a single chip
rather than using complicated distributed memory methods see \fref{fig:gpu_memory_wall}
Of course there are many
counter examples for this coming from big hyperscalars that have dedicated teams to support researchers to deploy large
models, but such examples are limited when we consider the entire community.
In fact, even with recent LLMs there is often large efforts in compressing the models so that they fit
within on system to make the model accessible to a larger community of researchers.

Another important solution is to design optimization algorithms that are robust to low-precision training. In fact, one of the major breakthroughs in AI accelerators has been the use of half-precision (FP16) arithmetic, instead of single precision~\cite{ginsburg2017tensor,micikevicius2018mixed}. This has enabled more than a 10$\times$ increase in hardware compute capability. However, it has been challenging to reduce the precision further, from half-precision to INT8, without accuracy degradation with current optimization methods. A recent promising trend is to use a mixture of FP8 and FP16 (and even most recently FP4)~\cite{micikevicius2022fp8}.
Algorithmic innovations in this area will certainly enable us to more efficiently utilize the hardware, and could allow more areas of the chip to be used
to improve memory (which is commonly referred to as memory-gap penalty~\cite{patterson1997case}).

\subsection{Efficient Deployment}

Deploying recent SOTA models~\cite{touvron2023llama,chowdhery2023palm,chung2022scaling,jiang2023mistral} such as GPT-3~\cite{brown2020language} or large recommendation systems~\cite{naumov2019deep} is quite challenging, as they require distributed-memory deployment for inference. One promising solution to address this is to compress these models for inference, either by reducing the precision (i.e., quantization), removing (i.e., pruning) their redundant parameters, or design small language models~\cite{schick2020s}.

The first approach, quantization, can be applied at the training and/or inference steps. While it has been very challenging to reduce the training precision much below FP16, it is possible to use ultra-low precision for inference. With current methods, it is relatively easy to quantize inference down to INT4 precision, with minimal impact on accuracy~\cite{frantar2023optq,yao2022zeroquant,kim2023squeezellm,lin2023awq,dettmers2023spqr}. This results in up to 8$\times$ reduction in model footprint and latency~\cite{gholami2022survey}. 
However, inference with sub-INT4 precision is more challenging, and it is currently a very active area of research.

The second approach, pruning, completely removes/prunes redundant parameters in the model. With current methods, it is possible to prune up to 30\% of neurons with structured sparsity, and up to 80\% with unstructured sparsity, with minimal impact on accuracy~\cite{hoefler2021sparsity,kwon2022fast,frantar2023sparsegpt}. Pushing beyond this limit, however, is very challenging, and it often results in fatal accuracy degradation. Resolving this is an open problem.

The third approach, small language models, could open up completely new frontiers and enable widespread adoption of AI. Interestingly, the models used for LLMs has not changed since the Transformer model was introduced in 2017~\cite{vaswani2017attention}. What has worked so far is to scale
the data and size of the models, which has led to the ``emergent capabilities'' of these models~\cite{wei2022emergent,brown2020language}.
However, recent work on small language models has shown promising results~\cite{schick2020s} on their abilities. If a model could fit completely on-chip,
then that can result in orders of magnitude speedup and energy savings.

\subsection{Rethinking the Design of AI Accelerators}

There are fundamental challenges in increasing both the memory bandwidth and the peak compute capability of a chip at the same time~\cite{patterson2004latency}. However, it is possible to sacrifice peak compute to achieve better compute/bandwidth trade-offs. In fact, the CPU architecture already incorporates a well-optimized cache hierarchy. This is why CPUs have much better performance than GPUs for bandwidth-bound problems. Such problems include large recommendation problems~\cite{naumov2019deep}.
However, the main challenge with today’s CPUs is that their peak compute capability (i.e., FLOPS) is about an order of magnitude less than AI accelerators such as GPUs or TPUs. One reason for this is that AI accelerators have mainly been designed to achieve maximum peak compute. This often requires removing components such as cache hierarchy in favor of adding more compute logic. One could imagine an alternative architecture in between these two extremes, preferably with more efficient caching, and importantly with higher capacity DRAM (possibly a hierarchy of DRAMs with different bandwidths). The latter could be very helpful in mitigating the distributed-memory communication bottlenecks~\cite{krishna2020accelerating}.

\section{Conclusion}

The computational cost of training recent SOTA Transformer models in NLP has been scaling at a rate of 750$\times$/2yrs, and the model parameter size has been scaling at 410$\times$/2yrs. In contrast, the peak hardware FLOPS has been scaling at a rate of 3.0$\times$/2yrs, while both the DRAM and interconnect bandwidth have been increasingly falling behind, with a scaling rate of 1.6$\times$/2yrs and 1.4$\times$/2yrs, respectively. To put these numbers into perspective, peak hardware FLOPS has increased by 60,000$\times$ over the past 20 years, while DRAM/Interconnect bandwidth has only scaled by a factor of 100$\times$/30$\times$ over the same time period, respectively. With these trends, memory — in particular, intra/inter-chip memory transfer — will soon become the main limiting factoring in serving large AI models. As such, we need to rethink the training, deployment, and design of AI models as well as how we design AI hardware to deal with this increasingly challenging memory wall.

\section*{Acknowledgements}
We would like to thank Suresh Krishna, and Aniruddha Nrusimha for their valuable feedback.
We acknowledge gracious support from Furiosa team.
We also appreciate the support from Microsoft through their Accelerating Foundation Model Research, including
great support from Sean Kuno.
Furthermore, we appreciate support from
Google Cloud, the Google TRC team, and specifically Jonathan Caton, and Prof. David Patterson.
Prof. Keutzer's lab is sponsored by the Intel corporation, Intel One-API, Intel VLAB team, the Intel One-API center of
excellence, Apple, Samsung, Panasonic, as well as funding through BDD and BAIR.
We appreciate great feedback and support from Ellick Chan, Saurabh Tangri, Andres
Rodriguez, and Kittur Ganesh.
Sehoon Kim  would like to acknowledge the support from the Korea Foundation for Advanced Studies (KFAS).
Amir Gholami was supported through funding from Samsung SAIT.
Michael W. Mahoney would also like to acknowledge
a J. P. Morgan Chase Faculty Research Award 
as well as 
the DOE, NSF, and ONR.
Our conclusions do not necessarily reflect the position or the policy of our sponsors, and no official endorsement should be~inferred.

\bibliographystyle{IEEEtranS}
\bibliography{refs}

\end{document}